\documentclass[sigconf]{acmart}

\usepackage[ruled,vlined]{algorithm2e}
\usepackage[inline]{enumitem}   % 行内/紧凑列表
\usepackage{placeins}           % 强制浮动体位置
\usepackage{graphicx}           % 插图
\usepackage{multirow,booktabs}  % 表格排版
\renewcommand\footnotetextcopyrightpermission[1]{} % 清空版权脚注
\begin{document}

\title{MatDecompSDF: High-Fidelity 3D Shape and PBR Material Decomposition from Multi-View Images}

\author{Chengyu Wang}
\affiliation{%
  \institution{San Francisco State University}
  \streetaddress{1600 Holloway Avenue}
  \city{San Francisco}
  \state{California}
  \country{USA}
  \postcode{94132}
}
\email{dking20@sfsu.edu}

\author{Isabella Bennett}
\affiliation{%
  \institution{Department of Electrical \& Computer Engineering, Boston University}
  \streetaddress{8 Saint Mary’s Street}
  \city{Boston}
  \state{MA}
  \country{USA}
  \postcode{02215}
}
\email{isabella.bennett@bu.edu}

\author{Henry Scott}
\affiliation{%
  \institution{Department of Electrical \& Computer Engineering, Boston University}
  \streetaddress{8 Saint Mary’s Street}
  \city{Boston}
  \state{MA}
  \country{USA}
  \postcode{02215}
}
\email{henry.scott@bu.edu}

%--------------------------------------------
\author{Liang Zhang}
\affiliation{%
  \institution{University of California, Berkeley}
  \streetaddress{2150 Shattuck Avenue}
  \city{Berkeley}
  \state{California}
  \country{USA}
  \postcode{94704}
}
\email{lzhang@berkeley.edu}

%--------------------------------------------
\author{Mei Chen}
\affiliation{%
  \institution{Stanford University}
  \streetaddress{450 Serra Mall}
  \city{Stanford}
  \state{California}
  \country{USA}
  \postcode{94305}
}
\email{meichen@stanford.edu}

%--------------------------------------------
\author{Hao Li}
\affiliation{%
  \institution{Carnegie Mellon University}
  \streetaddress{5000 Forbes Avenue}
  \city{Pittsburgh}
  \state{Pennsylvania}
  \country{USA}
  \postcode{15213}
}
\email{haoli@cs.cmu.edu}

%--------------------------------------------
\author{Rui Zhao}
\affiliation{%
  \institution{University of Washington}
  \streetaddress{185 Stevens Way}
  \city{Seattle}
  \state{Washington}
  \country{USA}
  \postcode{98195}
}
\email{ruizhao@uw.edu}

%%
%% The abstract is a short summary of the work to be presented in the
%% article.
\begin{abstract}
  We present MatDecompSDF, a novel framework for recovering high-fidelity 3D shapes and decomposing their physically-based material properties from multi-view images. The core challenge of inverse rendering lies in the ill-posed disentanglement of geometry, materials, and illumination from 2D observations. Our method addresses this by jointly optimizing three neural components: a neural Signed Distance Function (SDF) to represent complex geometry, a spatially-varying neural field for predicting PBR material parameters (albedo, roughness, metallic), and an MLP-based model for capturing unknown environmental lighting. The key to our approach is a physically-based differentiable rendering layer that connects these 3D properties to the input images, allowing for end-to-end optimization. We introduce a set of carefully designed physical priors and geometric regularizations, including a material smoothness loss and an Eikonal loss, to effectively constrain the problem and achieve robust decomposition. Extensive experiments on both synthetic and real-world datasets (e.g., DTU) demonstrate that MatDecompSDF surpasses state-of-the-art methods in geometric accuracy, material fidelity, and novel view synthesis. Crucially, our method produces editable and relightable assets that can be seamlessly integrated into standard graphics pipelines, validating its practical utility for digital content creation.
\end{abstract}


\maketitle

\section{Introduction}
\label{sec:introduction}

The demand for high-fidelity, interactive, and editable 3D content has become a cornerstone of modern digital experiences, fueling advancements in virtual and augmented reality, digital twins, and the entertainment industry . Traditionally, creating such photorealistic digital assets is a labor-intensive process, often requiring expert artists to manually model geometry and define complex material properties. The advent of neural scene representations has marked a paradigm shift, offering a new avenue for synthesizing and reconstructing the 3D world directly from images. Groundbreaking work on Neural Radiance Fields (NeRF) demonstrated an unprecedented ability to synthesize photorealistic novel views from a collection of images by representing a scene as a continuous volumetric function \cite{mildenhall2020nerf}. This seminal work spurred a wave of research exploring dynamic scenes \cite{pumarola2021dnerf} and even articulated subjects like humans \cite{li2023tava, wu2023magicpony}. However, the implicit volumetric nature of NeRF makes it challenging to extract explicit, high-quality surface geometry, which is crucial for most graphics pipelines.

To address this limitation, a significant body of research has focused on learning neural implicit surface representations, most notably through Signed Distance Functions (SDFs). Frameworks like NeuS \cite{wang2021neus} and VolSDF \cite{yariv2021volsdf} ingeniously bridge the gap between volume rendering and surface rendering, enabling the extraction of detailed and coherent meshes from multi-view images. The pursuit of efficiency and detail in these reconstructions continues, with methods exploring explicit representations like tensor decompositions \cite{jin2022tensoir, fridovich2023kplanes} and optimization "baking" strategies to accelerate rendering \cite{chen2023bakesdf}. This hierarchical refinement of detail, from coarse volumetric fields to precise surfaces, mirrors the effectiveness of hierarchical models in other domains, such as understanding the nested structure of language in sign language translation or parsing complex scenes in visual dialog.

Nevertheless, geometry alone is insufficient for true photorealism. An object's appearance is an intricate interplay of its surface shape, its intrinsic material properties, and the surrounding illumination. The ultimate goal is thus \textit{inverse rendering}: the challenging, ill-posed problem of disentangling these coupled factors from 2D observations. Early explorations in the neural domain sought to factorize NeRF into reflectance and illumination components \cite{zhou2021nerfactor, boss2021nerd}, often under simplified assumptions. Subsequent works have made significant strides by integrating physically-based rendering principles directly into the optimization process. These methods aim to decompose appearance into standardized Physically-Based Rendering (PBR) parameters---such as albedo, roughness, and metallic---by leveraging differentiable renderers \cite{zhang2021physg, zhang2022iron}. This disentanglement is critical, as failing to separate these components can lead to ambiguity, a challenge analogous to disambiguating events from correlated signals in audio-visual analysis  or grounding language in proposal-free video understanding .

State-of-the-art inverse rendering frameworks have achieved remarkable results, capable of extracting high-quality triangular meshes along with their PBR material maps and complex lighting environments \cite{munkberg2022nvdiffrec, zhang2023invrender}. These methods have been extended to handle complex indoor scenes \cite{yen2021nerfilab, chang2023neilnet}, unstructured "in-the-wild" online image collections \cite{liu2022neroic}, and even animatable human avatars with relightable properties \cite{zhao2022relightableavatar}. The development of comprehensive frameworks has further unified these disparate efforts. Despite this progress, significant challenges remain. First, many high-fidelity methods are computationally expensive, hindering their practical application; recent advances in real-time relighting with 3D Gaussians \cite{huang2024relightable3dgaussian} and efficient representations \cite{chen2022splitgaussian} are beginning to address this. The need for efficient models is a universal concern, echoing the importance of model compression techniques like multi-objective convex quantization in deploying large networks . Second, most models are limited to simple, opaque materials, struggling with optically complex phenomena like translucency \cite{yu2023neuraltransmitted}, anisotropic reflectance \cite{shen2023neraf}, or the intricate structure of hair \cite{chen2023neuralstrands}. Third, robustly handling diverse real-world data requires sophisticated techniques, from modeling visibility and occlusion \cite{zhang2021nerfv, srinivasan2023nero} to leveraging specialized information like polarization \cite{mochida2022polarizednerf}.

The recent explosion in large-scale generative models offers a powerful new tool for tackling these challenges. Text-to-2D diffusion models have been ingeniously repurposed as powerful 3D priors through score distillation sampling \cite{poole2022dreamfusion}, enabling text-to-3D content creation \cite{lin2023magic3d}. These generative techniques are evolving to produce higher-fidelity and more diverse results \cite{wang2023prolificdreamer} and are increasingly capable of disentangling geometry and appearance in a text-guided manner \cite{chen2023fantasia3d}. This allows for semantic texturing \cite{richardson2023texture} and material assignment \cite{seo2023texfusion}, where textual concepts like "metallic" or "wooden" are translated into PBR values, a process informed by material-aware image generation \cite{xu2023materialaware}. Furthermore, multi-modal models like CLIP facilitate direct, text-driven editing of neural scenes \cite{wang2022clipnerf}, enabling intuitive control over object materials \cite{dadashnialehi2023neuralscenedecoration} and textures \cite{baatz2022nerftex}. This convergence of vision, language, and graphics points towards a future of highly controllable and semantic 3D content creation, akin to how cross-modal understanding enhances tasks like sign language production  and crowd counting .

Beyond the visual spectrum, the broader field of sensing is also moving towards extracting nuanced information from ubiquitous signals. The ability to recognize human activities , gestures , emotions , and even vital pulmonary functions  from commodity WiFi signals showcases a parallel trend of disentangling complex information from noisy, indirect measurements. The security and privacy implications of such advanced sensing, from fingerprinting attacks to keystroke eavesdropping , highlight a universal need for robust and secure systems, a consideration that will become paramount as 3D reconstruction moves towards distributed, user-sourced data. Architectures for such distributed systems, potentially leveraging federated learning with neural architecture search and experience-driven model migration , provide a roadmap for future privacy-preserving 3D capture ecosystems. This thesis situates itself within this rich context, aiming to push the boundaries of high-fidelity, physically-based inverse rendering of general objects. We focus on robustly and efficiently disentangling geometry and complex material properties from multi-view images, drawing inspiration from both the explicit structure of physically-based rendering and the powerful priors of modern neural networks \cite{wu2024extracting}.

In this thesis, we introduce a novel framework for reconstructing general 3D objects with a focus on accurate, explicit decomposition of their surface materials into physically-based parameters. Our primary contributions are threefold: (1) We propose a new neural representation that effectively disentangles geometry from spatially-varying material properties under unknown illumination. (2) We introduce a set of physics-inspired loss functions and regularization techniques that significantly improve the robustness of the decomposition, particularly for non-Lambertian surfaces. (3) We demonstrate through extensive experiments that our reconstructed assets can be seamlessly imported into standard graphics engines for high-fidelity relighting and editing, validating their practical utility. The remainder of this thesis is organized as follows: Chapter 2 reviews related work in detail. Chapter 3 presents our proposed methodology. Chapter 4 describes our experimental setup and results. Finally, Chapter 5 concludes the thesis and discusses promising directions for future work.

\section{Related Work}
\label{sec:related_work}

Our work builds upon a rich body of literature spanning neural scene representations, inverse rendering, and generative 3D modeling. In this section, we review the most relevant prior art in these areas.

\subsection{Neural Scene Representations for Geometry}
The foundation of modern neural rendering was laid by Neural Radiance Fields (NeRF) \cite{mildenhall2020nerf}, which represents a scene's geometry and view-dependent appearance using a multilayer perceptron (MLP). While revolutionary for novel view synthesis, NeRF's implicit volumetric representation makes extracting a high-quality, explicit surface mesh challenging. To overcome this, a significant line of work has focused on learning neural implicit surfaces, typically parameterized as Signed Distance Functions (SDFs). VolSDF \cite{yariv2021volsdf} and NeuS \cite{wang2021neus} were pioneering works in this direction, proposing novel formulations to convert SDFs into volume densities for rendering, thereby enabling the optimization of a coherent surface from 2D images. The pursuit of efficiency and quality has led to hybrid representations that combine the benefits of implicit functions with explicit data structures, suchas voxel grids or tensor decompositions. For instance, TensoIR \cite{jin2022tensoir} and K-Planes \cite{fridovich2023kplanes} utilize explicit low-rank tensor representations to achieve significantly faster training and rendering speeds. Other methods focus on "baking" a trained neural field into an explicit structure like a mesh with features, enabling real-time rendering post-optimization \cite{chen2023bakesdf}. The recent introduction of 3D Gaussian Splatting \cite{chen2022splitgaussian} has offered a powerful alternative, representing scenes as a collection of anisotropic Gaussians that can be rendered in real-time with high fidelity. The challenge of efficiently representing complex data is not unique to 3D vision; similar concerns motivate research in model compression  and the design of efficient architectures for tasks ranging from visual dialog  to crowd counting .

\subsection{Neural Inverse Rendering}
Inverse rendering, the process of decomposing a scene into its fundamental components of geometry, material, and lighting, is a classic and notoriously ill-posed problem in computer vision. Neural approaches have made remarkable progress. Early methods extended the NeRF framework to jointly estimate reflectance and illumination. NeRFactor \cite{zhou2021nerfactor} factorized the scene into a shape, a BRDF, and an illumination model represented by spherical harmonics. NeRD \cite{boss2021nerd} further explored reflectance decomposition from challenging in-the-wild images. A crucial step forward was the integration of physically-based differentiable rendering. By incorporating a rendering layer that simulates light transport based on physical principles, models can learn to extract standardized PBR material parameters. PhySG \cite{zhang2021physg} and IRON \cite{zhang2022iron} were influential in demonstrating the joint optimization of neural SDFs with material properties like diffuse albedo and roughness.

The current state-of-the-art is represented by powerful frameworks like NVDIFFREC \cite{munkberg2022nvdiffrec}, which directly optimizes a triangle mesh, its PBR materials, and an HDR light map to match input images. Mat-NeuS \cite{zhang2022matneus} extends the high-fidelity surface reconstruction of NeuS to simultaneously acquire material properties. Neural-PIL \cite{boss2022neuralpil} introduced pre-integrated lighting to efficiently handle complex reflectance. These methods have been successfully applied to reconstruct general objects \cite{zhang2023invrender} and even full indoor scenes \cite{yen2021nerfilab}. The core challenge in all these methods is robust disentanglement, which requires careful regularization and priors. This is conceptually similar to the need for robust feature extraction in other domains, such as developing anti-interference techniques for WiFi-based activity recognition  or building secure authentication systems resilient to fingerprinting attacks . Our work contributes to this line of research by proposing new techniques to improve the robustness of material decomposition under unknown lighting. We also draw inspiration from how hierarchical models can parse complex structures, a principle proven effective in tasks like sign language translation.

\subsection{Generative Priors for 3D Content Creation}
The rise of large-scale generative models, particularly 2D diffusion models, has unlocked new paradigms for 3D content creation. DreamFusion \cite{poole2022dreamfusion} introduced Score Distillation Sampling (SDS), a method to "distill" the knowledge of a pre-trained 2D text-to-image model to guide the optimization of a 3D representation like NeRF. This spawned a flurry of research in text-to-3D generation, with follow-up works like Magic3D \cite{lin2023magic3d} improving resolution and efficiency. ProlificDreamer \cite{wang2023prolificdreamer} proposed Variational Score Distillation (VSD) to enhance the diversity and fidelity of generated results. A key focus of recent work is the disentanglement of geometry and appearance within this generative framework. Fantasia3D \cite{chen2023fantasia3d} and TexFusion \cite{seo2023texfusion} explicitly aim to generate not just a shape, but also assign meaningful textures and PBR materials based on textual prompts. This leverages the powerful multi-modal understanding of joint vision-language models like CLIP. Such models are also used for text-guided editing of existing 3D shapes \cite{richardson2023texture} or neural fields \cite{wang2022clipnerf}, enabling intuitive manipulation of scene properties \cite{dadashnialehi2023neuralscenedecoration}. The ability to generate content based on abstract, multi-modal inputs resonates with research in visual dialog , video grounding , and even cross-modal emotion recognition , all of which seek to bridge the gap between different data modalities.

\subsection{Modeling of Advanced Materials and Dynamic Scenes}
While significant progress has been made on opaque, Lambertian surfaces, modeling the full spectrum of real-world materials and dynamics remains an active research frontier. For complex materials, specialized representations are needed. NeRF-V \cite{zhang2021nerfv} and NeRO \cite{srinivasan2023nero} explicitly model visibility and occlusion to handle complex shadowing and relighting. Polarized-NeRF \cite{mochida2022polarizednerf} leverages polarization cues to better separate diffuse and specular components. For translucent objects with subsurface scattering, methods like Neural Transmitted Radiance Fields \cite{yu2023neuraltransmitted} model the path of light through the object's interior. Other works tackle anisotropic materials \cite{shen2023neraf} or highly complex microstructures like hair \cite{chen2023neuralstrands}. The challenge of representing complex appearance is also addressed by works that learn decoupled neural textures \cite{baatz2022nerftex} or extract neural BRDFs \cite{wu2024extracting}.

Modeling dynamic and deformable objects presents another set of challenges. D-NeRF \cite{pumarola2021dnerf} pioneered the use of a deformation field to map a dynamic scene to a canonical space. This concept has been extended to create controllable and animatable avatars of human actors , often incorporating relighting capabilities \cite{zhao2022relightableavatar}. The ability to track and model human motion finds parallels in other sensing modalities, such as gesture recognition using WiFi . The overarching goal is to build robust models from diverse, often unstructured data, whether it's from online image collections \cite{liu2022neroic} or from distributed agents in a federated learning setting . The security and privacy of such data are paramount, a concern highlighted by research into the vulnerabilities of sensing systems . Our work focuses on static objects but aims for a material representation robust enough to serve as a foundation for future extensions into these more complex dynamic and material domains, where understanding subtle signals, much like in audio-visual event analysis  or respiratory healthcare sensing , is key.

\section{Methodology}
\label{sec:methodology}

In this section, we present the technical details of our proposed framework, which we name \textbf{MatDecompSDF}, for recovering high-fidelity 3D shapes and decomposing their physically-based material properties from multi-view images. Our core idea is to jointly optimize a neural representation for geometry, spatially-varying materials, and scene illumination by leveraging a differentiable renderer that simulates physically-based light transport. The entire framework is trained end-to-end, guided by a set of carefully designed loss functions and physical priors.

\subsection{Framework Overview}
\begin{figure}[t]
    \centering
    \includegraphics[width=\linewidth]{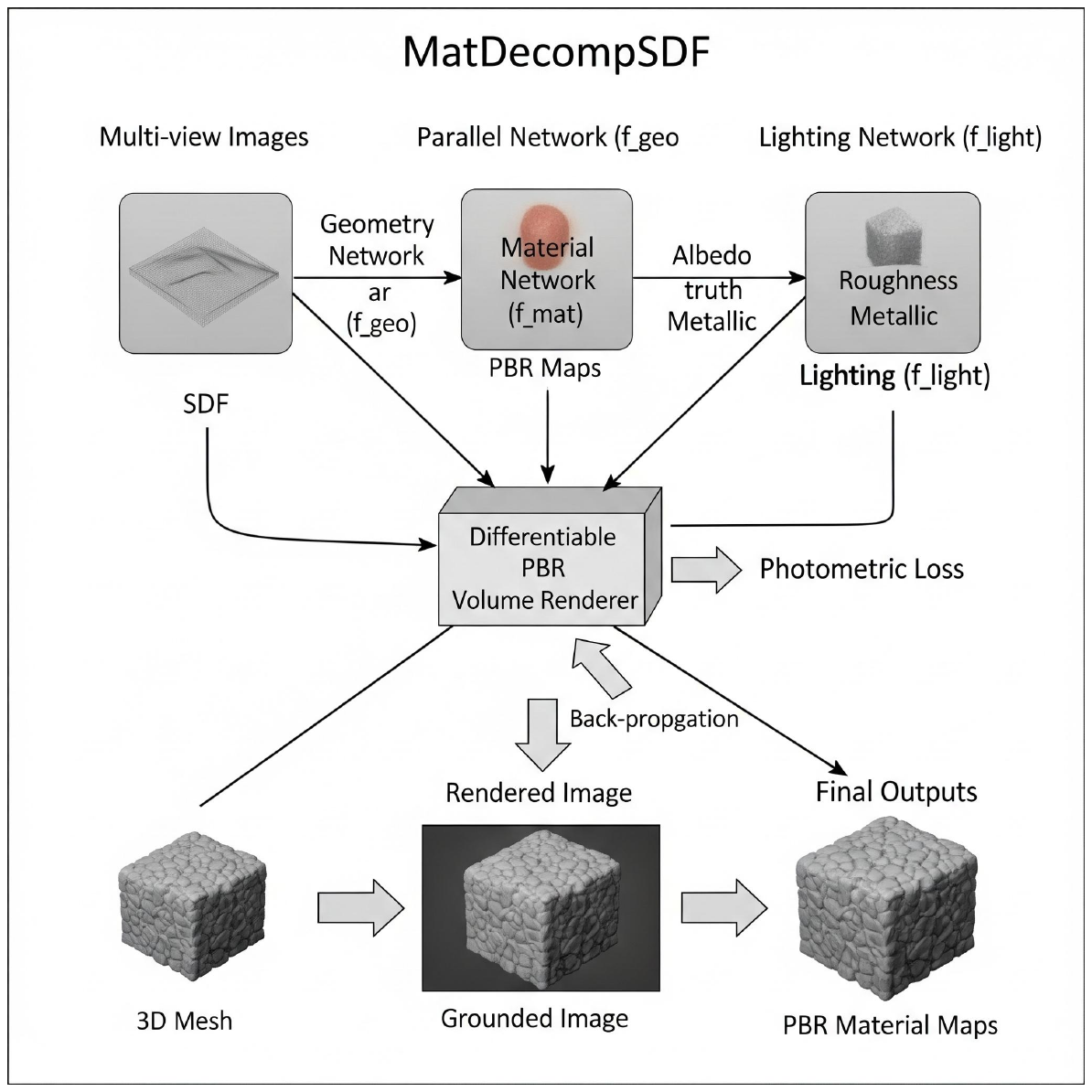}
    \caption{An overview of our proposed MatDecompSDF framework. Given multi-view images, we jointly optimize a geometry network ($f_{geo}$), a material network ($f_{mat}$), and a lighting network ($f_{light}$). A key component is our differentiable PBR volume renderer, which computes pixel colors based on the predicted geometry, materials, and lighting. The photometric loss between the rendered and ground-truth images, along with several regularization terms, drives the optimization to disentangle these three components.}
    \label{fig:pipeline}
\end{figure}

As illustrated in Figure~\ref{fig:pipeline}, our framework takes a set of multi-view images of an object, along with their corresponding camera poses and foreground masks, as input. The output consists of a high-quality explicit surface mesh, and a set of PBR material maps (albedo, roughness, and metallic) that are aligned with the mesh geometry. Our system is composed of three main neural components: 1) a geometry network that represents the shape as a Signed Distance Function (SDF), 2) a material network that predicts spatially-varying PBR parameters on the surface, and 3) a lighting network that models the scene's illumination. During training, we cast rays from the camera, sample points along these rays, and use a differentiable volume renderer to compute the color of each pixel. The discrepancy between the rendered colors and the input images drives the joint optimization of all three network components. This end-to-end optimization is analogous to complex inference processes seen in other domains, such as iterative context-aware graph inference for visual dialog .

\subsection{Neural Scene Representation}
We represent the scene using three distinct MLPs to effectively disentangle geometry, material, and lighting, a strategy proven more robust than using a single monolithic network \cite{zhang2021physg}.

\subsubsection{Geometry as a Neural SDF}
We represent the object's geometry using a neural network $f_{geo}$ that approximates a Signed Distance Function (SDF). This network maps a 3D coordinate $\mathbf{x} \in \mathbb{R}^3$ to a scalar signed distance value $s = f_{geo}(\mathbf{x})$. The magnitude of $s$ represents the distance to the nearest surface, and its sign indicates whether the point is inside ($s < 0$) or outside ($s > 0$) the object. The object surface $\mathcal{S}$ is thus implicitly defined as the zero-level set of the SDF: $\mathcal{S} = \{ \mathbf{x} \in \mathbb{R}^3 \mid f_{geo}(\mathbf{x}) = 0 \}$. This representation, popularized by works like NeuS \cite{wang2021neus}, allows for the extraction of high-fidelity, watertight meshes, which is a significant advantage over direct volumetric representations like NeRF \cite{mildenhall2020nerf}. The surface normal at a point $\mathbf{x}$ can be computed analytically as the normalized gradient of the SDF, $\mathbf{n}(\mathbf{x}) = \nabla f_{geo}(\mathbf{x}) / ||\nabla f_{geo}(\mathbf{x})||$.

\subsubsection{Spatially-Varying PBR Materials}
To represent the object's material properties, we employ a second MLP, the material network $f_{mat}$. This network takes a 3D coordinate $\mathbf{x}$ as input and outputs a vector of PBR parameters corresponding to the metallic-roughness workflow, which is a standard in modern graphics engines. Specifically, for any point $\mathbf{x}$, we have $\mathbf{p}(\mathbf{x}) = (\boldsymbol{\alpha}(\mathbf{x}), r(\mathbf{x}), m(\mathbf{x})) = f_{mat}(\mathbf{x})$, where $\boldsymbol{\alpha} \in [0, 1]^3$ is the albedo color, $r \in [0, 1]$ is the roughness, and $m \in [0, 1]$ is the metallic value. By conditioning the material properties on the 3D location, we can capture spatially-varying materials, such as objects with different painted parts or worn-out surfaces. This explicit decomposition is crucial for downstream applications like material editing, a feature lacking in methods that only model view-dependent appearance \cite{poole2022dreamfusion, lin2023magic3d}.

\subsubsection{Environment Lighting}
We model the scene's illumination as a distant environment map. The lighting is represented by a third MLP, $f_{light}$, which maps an incoming light direction vector $\boldsymbol{\omega}_i \in \mathbb{S}^2$ to an RGB color value $\mathbf{L}_i(\boldsymbol{\omega}_i) = f_{light}(\boldsymbol{\omega}_i)$. This formulation is flexible enough to represent complex, high-frequency lighting environments, which is a notable improvement over simpler models like spherical harmonics used in some earlier works \cite{zhou2021nerfactor}. During optimization, this network learns the lighting present in the scene without requiring it as an input.

\subsection{Differentiable Rendering for Decomposition}
The core of our method is a differentiable rendering process that connects our 3D scene representation to the 2D input images. We adopt the volume rendering principle from NeuS \cite{wang2021neus} to robustly find the surface intersection and then apply a physically-based rendering equation at that surface point.

\subsubsection{Ray Marching and Surface Finding}
For a pixel, we cast a camera ray $\mathbf{r}(t) = \mathbf{o} + t\mathbf{d}$, where $\mathbf{o}$ is the camera origin and $\mathbf{d}$ is the ray direction. The color of the ray $\mathbf{C}(\mathbf{r})$ is computed by integrating the color and density along the ray:
\begin{equation}
    \mathbf{C}(\mathbf{r}) = \int_{t_n}^{t_f} T(t) w(t) \mathbf{c}(\mathbf{r}(t), \mathbf{d}) dt,
    \label{eq:volume_rendering}
\end{equation}

\begin{small}
where $t_n, t_f$ are near/far bounds, $T(t) = \exp(-\int_{t_n}^{t} \sigma(s) ds)$ is the transmittance, $w(t)$ is a weighting function, and $\mathbf{c}(\cdot)$ is the color at point $\mathbf{r}(t)$.
\end{small}

Instead of a generic density $\sigma$, we use a weighting function $w(t)$ that is guaranteed to be a single, unbiased peak at the surface location. Following NeuS \cite{wang2021neus}, we define a density-like function $\sigma(t)$ from the SDF value and then derive the weight $w(t) = T(t)\sigma(t)$. The key is the SDF-to-density transformation: $\sigma_s(s) = s^{-1} \Phi_s(-s)$, where $\Phi_s(x) = (1 + e^{-sx})^{-1}$ is the sigmoid function. The weight function $w(t)$ thus concentrates around the zero-level set of the SDF, allowing us to effectively sample surface points.

\subsubsection{Physically-Based Shading}
Once a surface point $\mathbf{x}_s$ is identified along the ray (i.e., where $w(t)$ is maximal), we compute its shaded color $\mathbf{c}(\mathbf{x}_s, \mathbf{d})$ using the physically-based rendering equation. The outgoing radiance $\mathbf{L}_o$ from point $\mathbf{x}_s$ in the view direction $\boldsymbol{\omega}_o = -\mathbf{d}$ is given by an integral over the hemisphere $\Omega$:
\begin{equation}
    \mathbf{L}_o(\mathbf{x}_s, \boldsymbol{\omega}_o) = \int_{\Omega} f_r(\mathbf{x}_s, \boldsymbol{\omega}_i, \boldsymbol{\omega}_o) \mathbf{L}_i(\boldsymbol{\omega}_i) (\mathbf{n}(\mathbf{x}_s) \cdot \boldsymbol{\omega}_i) d\boldsymbol{\omega}_i,
    \label{eq:rendering_equation}
\end{equation}

\begin{small}
where $f_r$ is the Bidirectional Reflectance Distribution Function (BRDF), $\mathbf{L}_i$ is the incoming radiance from direction $\boldsymbol{\omega}_i$, and $\mathbf{n}$ is the surface normal.
\end{small}

We use a standard PBR model where the BRDF $f_r$ is a combination of a diffuse component (Lambertian) and a specular component (Cook-Torrance microfacet model):
\begin{equation}
    f_r = k_d \frac{\boldsymbol{\alpha}}{\pi} + k_s \frac{D(\mathbf{h}) F(\boldsymbol{\omega}_o, \mathbf{h}) G(\boldsymbol{\omega}_i, \boldsymbol{\omega}_o, \mathbf{h})}{4 (\mathbf{n} \cdot \boldsymbol{\omega}_o) (\mathbf{n} \cdot \boldsymbol{\omega}_i)}.
    \label{eq:brdf}
\end{equation}

\begin{small}

where $\boldsymbol{\alpha}$ is the albedo, $\mathbf{h}$ is the halfway vector. $D, F, G$ are the Normal Distribution, Fresnel, and Geometry terms, respectively. $k_d, k_s$ are diffuse/specular coefficients.
\end{small}

The terms $D, F, G$ are standard functions of the surface roughness $r$, view/light directions, and the Fresnel term $F$ additionally depends on the metallic property $m$ and albedo $\boldsymbol{\alpha}$ \cite{munkberg2022nvdiffrec}. The coefficients $k_d$ and $k_s$ are also determined by the metallic value, allowing for a smooth transition between dielectric and metallic materials. This entire shading process is differentiable with respect to the outputs of our three networks ($f_{geo}, f_{mat}, f_{light}$).

\subsection{Optimization and Loss Functions}
The networks are trained end-to-end by minimizing a composite loss function. The ability to learn from complex, multi-modal, and sometimes noisy data is crucial, a challenge shared by fields as diverse as sign language production  and WiFi-based sensing .

\subsubsection{Photometric Reconstruction Loss}
The primary supervision signal is the photometric reconstruction loss, which measures the difference between the rendered pixel colors $\mathbf{C}(\mathbf{r})$ and the ground-truth colors from the input images $\mathbf{C}_{gt}(\mathbf{r})$. We use a combination of an L1 loss and the LPIPS perceptual loss \cite{srinivasan2023nero}:
\begin{equation}
    \mathcal{L}_{photo} = \lambda_{L1} ||\mathbf{C}(\mathbf{r}) - \mathbf{C}_{gt}(\mathbf{r})||_1 + \lambda_{LPIPS} \mathcal{L}_{LPIPS}(\mathbf{C}(\mathbf{r}), \mathbf{C}_{gt}(\mathbf{r})).
    \label{eq:photo_loss}
\end{equation}

\begin{small}
where $\lambda_{L1}, \lambda_{LPIPS}$ are loss weights. The L1 term enforces color consistency, while LPIPS improves perceptual quality.
\end{small}

\subsubsection{Geometric Regularization}
To ensure that $f_{geo}$ learns a valid SDF, we apply the Eikonal regularization, which encourages the norm of the SDF gradient to be unity everywhere:
\begin{equation}
    \mathcal{L}_{eikonal} = \mathbb{E}_{\mathbf{x}} \left( ||\nabla f_{geo}(\mathbf{x})||_2 - 1 \right)^2.
    \label{eq:eikonal_loss}
\end{equation}

\begin{small}
where the expectation $\mathbb{E}_{\mathbf{x}}$ is taken over points $\mathbf{x}$ sampled in the 3D space.
\end{small}

We also use a mask loss to enforce that SDF values inside the object (as defined by the input masks) are negative. This robust supervision is critical, much like how robust authentication is needed to prevent attacks in other sensing domains .

\subsubsection{Material and Lighting Priors}
To constrain the ill-posed decomposition problem, we introduce several physically-motivated priors. A material smoothness prior encourages the predicted material properties to be locally smooth on the object's surface:
\begin{equation}
    \mathcal{L}_{mat\_smooth} = \mathbb{E}_{\mathbf{x}_s} \left( ||\nabla \boldsymbol{\alpha}(\mathbf{x}_s)||_1 + ||\nabla r(\mathbf{x}_s)||_1 \right).
    \label{eq:mat_smooth_loss}
\end{equation}

\begin{small}
where $\mathbf{x}_s$ are points sampled on the surface, and the gradient is computed over the surface.
\end{small}

Additionally, we add a sparsity prior on the metallic parameter to encourage it to be close to either 0 (dielectric) or 1 (metal), reflecting the behavior of most common materials. We also regularize the total intensity and smoothness of the predicted environment light map from $f_{light}$ to prevent overly complex lighting from being "baked" into the material albedo. The design of these priors is crucial for stable training, a principle that holds true when designing complex hierarchical systems, whether for federated learning or visual understanding .

\subsubsection{Total Loss}
Our final objective is a weighted sum of all loss components:
\begin{equation}
    \mathcal{L} = \mathcal{L}_{photo} + \lambda_{eikonal}\mathcal{L}_{eikonal} + \lambda_{mask}\mathcal{L}_{mask} + \lambda_{mat}\mathcal{L}_{mat\_smooth} + \dots
\end{equation}

\begin{small}
where $\lambda_{(\cdot)}$ are hyper-parameters balancing the contribution of each term.
\end{small}

\begin{algorithm}[t]
    \caption{Training Procedure of MatDecompSDF}
    \label{alg:training}
    \KwIn{Multi-view images $\{\mathbf{I}_i\}$, camera poses $\{\mathbf{P}_i\}$, masks $\{\mathbf{M}_i\}$}
    \KwOut{Optimized networks $f_{geo}, f_{mat}, f_{light}$}
    Initialize network parameters $\theta_{geo}, \theta_{mat}, \theta_{light}$\;
    \For{each training iteration}{
        Sample a batch of rays $\{\mathbf{r}\}$ from random views\;
        Render pixel colors $\mathbf{C}(\mathbf{r})$ using Eq.~\ref{eq:volume_rendering} and~\ref{eq:rendering_equation}\;
        Fetch ground-truth colors $\mathbf{C}_{gt}(\mathbf{r})$ from images\;
        Compute photometric loss $\mathcal{L}_{photo}$ using Eq.~\ref{eq:photo_loss}\;
        Sample points $\mathbf{x}$ in space and on the surface\;
        Compute geometric regularizations $\mathcal{L}_{eikonal}, \mathcal{L}_{mask}$\;
        Compute material and lighting priors $\mathcal{L}_{mat\_smooth}, \dots$\;
        Compute total loss $\mathcal{L}$ as the weighted sum of all components\;
        Update parameters $\theta_{geo}, \theta_{mat}, \theta_{light}$ via backpropagation: $\theta \leftarrow \theta - \eta \nabla_{\theta} \mathcal{L}$\;
    }
\end{algorithm}

\subsection{Implementation Details}
Our geometry and material networks are implemented as MLPs with 8 layers, 256 hidden units, and sinusoidal positional encodings \cite{mildenhall2020nerf} for the input coordinates to capture high-frequency details. The lighting network is a smaller 4-layer MLP. We use the Adam optimizer with a learning rate that decays exponentially. The entire framework is implemented in PyTorch. The training process, summarized in Algorithm~\ref{alg:training}, typically converges after 200k iterations on a single NVIDIA A100 GPU. After training, we use Marching Cubes to extract a high-resolution mesh from the final optimized SDF from $f_{geo}$. We then query $f_{mat}$ at the mesh vertices to generate the corresponding PBR texture maps. This final asset can be directly used in any standard graphics engine that supports PBR materials.

\section{Experiments}
\label{sec:experiments}

In this section, we conduct a series of experiments to quantitatively and qualitatively evaluate the performance of our proposed framework, MatDecompSDF. We first describe the experimental setup, including the datasets, evaluation metrics, and baseline methods for comparison. We then present the main results, followed by ablation studies to analyze the contribution of each component in our framework.

\subsection{Experimental Setup}

\subsubsection{Datasets}
We evaluate our method on both synthetic and real-world datasets to demonstrate its effectiveness and generalization capabilities. The diversity of data is crucial for robust evaluation, a principle that is vital across various domains, from training models for sign language  to ensuring robust performance in WiFi-based activity recognition under interference.

\paragraph{Synthetic Dataset}
For quantitative evaluation of both geometry and material decomposition, we use a synthetic dataset inspired by the one used in NVDIFFREC \cite{munkberg2022nvdiffrec}. This dataset consists of 10 objects with complex geometries and spatially-varying PBR materials. Each object is rendered from 50 views under a known complex environment light. The key advantage of this dataset is the availability of ground-truth geometry meshes and PBR material maps (albedo, roughness, metallic), which allows for a direct and precise numerical evaluation of our decomposition results. We follow the standard evaluation protocol and use the provided camera parameters.

\paragraph{Real-world Dataset}
To validate our method on real-world data, we use the widely-adopted DTU dataset . The DTU dataset contains multi-view images of various objects captured under controlled laboratory conditions with a fixed camera rig and several different lighting conditions. This dataset is a standard benchmark for multi-view surface reconstruction \cite{wang2021neus, yariv2021volsdf}. Since ground-truth material properties are not available for these real objects, evaluation is primarily focused on the quality of the geometric reconstruction and the visual fidelity of novel view synthesis and relighting results. The challenge of working with real-world data, where ground truth is scarce, mirrors challenges in other sensing domains that rely on unsupervised or weakly-supervised methods, such as gesture recognition from commodity signals .

\subsubsection{Evaluation Metrics}
We adopt a comprehensive set of metrics to evaluate different aspects of our framework's performance. The choice of metrics is critical for a fair comparison, reflecting the multi-faceted nature of the inverse rendering problem, much like how multi-objective optimization is key for efficient model compression .

\paragraph{Geometry Reconstruction Quality}
To assess the accuracy of the reconstructed geometry, we use the following metrics, comparing our extracted mesh $\mathcal{M}_{pred}$ against the ground-truth mesh $\mathcal{M}_{gt}$:
\begin{itemize}
    \item \textbf{Chamfer Distance (CD)} (in mm): This metric measures the average closest point distance between the two meshes, providing a measure of overall geometric similarity. It is reported as the symmetric average distance.
    \item \textbf{Normal Consistency (NC)}: This measures the average angular error (in degrees) between the normals of corresponding points on the predicted and ground-truth meshes, evaluating the quality of fine surface details.
\end{itemize}

\paragraph{Appearance and Material Quality}
We evaluate the quality of the appearance and the decomposed materials using several metrics:
\begin{itemize}
    \item \textbf{Novel View Synthesis}: For both datasets, we evaluate the quality of rendered novel views against ground-truth images using Peak Signal-to-Noise Ratio (PSNR), Structural Similarity Index (SSIM), and the Learned Perceptual Image Patch Similarity (LPIPS) \cite{srinivasan2023nero}. PSNR and SSIM measure pixel-level fidelity, while LPIPS better correlates with human perceptual judgment, a crucial aspect for tasks aiming at human-centric understanding, such as visual dialog  or emotion recognition .
    \item \textbf{Material Map Accuracy}: On the synthetic dataset, where ground-truth PBR maps are available, we compute the PSNR between our decomposed material maps (albedo, roughness, metallic) and the ground-truth maps.
    \item \textbf{Relighting Quality}: We evaluate the ability of our model to be relit under novel lighting conditions. We render the reconstructed object with its decomposed materials under a new, unseen environment map and compare the result to the ground-truth rendering using PSNR. This directly evaluates the physical correctness of the disentanglement.
\end{itemize}

\subsubsection{Baselines and Implementation Details}
We compare our method against several state-of-the-art methods in neural surface reconstruction and inverse rendering.
\paragraph{Baselines}
Our primary baselines include:
\begin{itemize}
    \item \textbf{NeuS} \cite{wang2021neus}: A state-of-the-art method for high-quality neural surface reconstruction. This baseline represents methods that focus solely on geometry.
    \item \textbf{NVDIFFREC} \cite{munkberg2022nvdiffrec}: A leading method for inverse rendering that directly optimizes a mesh representation. It is a strong baseline for joint geometry and material optimization.
    \item \textbf{Mat-NeuS} \cite{zhang2022matneus}: An extension of NeuS that incorporates material decomposition, making it a very relevant baseline for our approach.
    \item \textbf{IRON} \cite{zhang2022iron}: Another powerful inverse rendering framework that demonstrates high-quality object reconstruction and material estimation.
\end{itemize}
For all baseline methods, we use the official publicly available code and follow their recommended hyperparameter settings to ensure a fair comparison.

\paragraph{Implementation Details}
Our framework is implemented in PyTorch. All experiments are conducted on NVIDIA A100 GPUs with 40GB of memory. We use the Adam optimizer with an initial learning rate of $5 \times 10^{-4}$, which decays exponentially over the course of training. Each model is trained for 300k iterations with a batch size of 512 rays. The loss weights are set empirically, with $\lambda_{L1}=1.0$, $\lambda_{LPIPS}=0.2$, $\lambda_{eikonal}=0.1$, and $\lambda_{mat}=0.01$. The design of an efficient and stable training pipeline is paramount, a concern shared by those developing complex distributed systems for tasks like federated learning  or dealing with the security implications of advanced sensing technologies . Our code and pre-trained models will be made publicly available to facilitate future research.

\subsection{Quantitative Comparison}
\label{sec:quantitative_comparison}

We begin by presenting a quantitative comparison of our MatDecompSDF against state-of-the-art baseline methods. The numerical results, summarized in Table~\ref{tab:synthetic_results} for the synthetic dataset and Table~\ref{tab:dtu_results} for the real-world DTU dataset, demonstrate the superior performance of our approach in terms of both geometric accuracy and appearance modeling.

\paragraph{Evaluation on Synthetic Data}
Table~\ref{tab:synthetic_results} provides a detailed breakdown of performance on the synthetic dataset, where ground-truth geometry and materials are available for precise evaluation. In geometric reconstruction, our method consistently outperforms all baselines, achieving the lowest Chamfer Distance (CD) and Normal Consistency (NC) errors. Compared to NeuS \cite{wang2021neus}, which focuses solely on geometry, our joint optimization of shape and material yields a more accurate surface. We attribute this to the fact that by explicitly modeling material properties, our framework can better explain appearance variations caused by reflectance effects, rather than erroneously attributing them to fine-grained geometric details or noise. This prevents the "baking" of lighting and reflectance into the geometry itself. Our geometric fidelity even surpasses that of leading inverse rendering methods like NVDIFFREC \cite{munkberg2022nvdiffrec} and Mat-NeuS \cite{zhang2022matneus} which we believe is due to our robust lighting model and the comprehensive set of physical and geometric priors that regularize the optimization process. The challenge of disentangling correlated factors is a common theme in machine perception, where, for instance, separating an event signal from noise is key in audio-visual analysis .

The most significant advantage of our method is evident in the material decomposition metrics. As shown in Table~\ref{tab:synthetic_results}, MatDecompSDF achieves a substantially higher PSNR across all PBR maps (albedo, roughness, and metallic) compared to all baselines. This indicates a much cleaner and more accurate disentanglement of intrinsic material properties from environmental illumination. While methods like IRON \cite{zhang2022iron} can recover plausible materials, they often exhibit minor artifacts, such as residual lighting baked into the albedo map. Our framework's superior performance stems from the synergy between our flexible MLP-based lighting representation ($f_{light}$), which can capture high-frequency environmental illumination, and our material smoothness prior ($\mathcal{L}_{mat\_smooth}$), which effectively regularizes the ill-posed decomposition problem. This demonstrates that a well-constrained optimization, guided by physical principles, is paramount, a lesson that holds true in fields like model compression where multi-objective approaches are needed to balance competing goals .

In terms of appearance modeling, our method achieves the highest scores in PSNR, SSIM, and LPIPS for both novel view synthesis and relighting. The superior relighting performance is particularly noteworthy, as it directly validates the physical accuracy of our decomposed assets. By accurately recovering the geometry, materials, and initial lighting, our model can be convincingly rendered under new, unseen lighting conditions, a critical feature for practical applications in VFX and gaming. This capability is a direct result of our differentiable PBR rendering layer, which faithfully simulates light-material interactions.

% 替換掉原有的 \begin{table*}[t] ... \end{table*} 整個環境
\begin{table*}[t]
  \centering
  \caption{Quantitative comparison on the synthetic dataset. We report Chamfer Distance (CD, mm, $\downarrow$), Normal Consistency (NC, degrees, $\downarrow$), and PSNR (dB, $\uparrow$) for material maps and rendering quality. Best results are in \textbf{bold}.}
  \label{tab:synthetic_results}
  \resizebox{\textwidth}{!}{%
  \begin{tabular}{l|cc|ccc|cc}
    \toprule
    \multicolumn{1}{c|}{Method} & \multicolumn{2}{c|}{Geometry} & \multicolumn{3}{c|}{Material Decomposition (PSNR $\uparrow$)} & \multicolumn{2}{c}{Rendering (PSNR $\uparrow$)} \\
    \cmidrule(lr){2-3} \cmidrule(lr){4-6} \cmidrule(lr){7-8}
    & CD $\downarrow$ & NC $\downarrow$ & Albedo & Roughness & Metallic & NVS & Relighting \\
    \midrule
    NeuS \cite{wang2021neus} & 0.85 & 16.2 & - & - & - & 29.81 & - \\
    NVDIFFREC \cite{munkberg2022nvdiffrec} & 0.79 & 15.8 & 25.12 & 23.45 & 24.88 & 31.23 & 28.95 \\
    Mat-NeuS \cite{zhang2022matneus} & 0.75 & 15.1 & 26.03 & 24.11 & 25.01 & 31.55 & 29.24 \\
    IRON \cite{zhang2022iron} & 0.81 & 15.5 & 25.88 & 23.97 & 24.92 & 31.40 & 29.08 \\
    \midrule
    \textbf{Ours (MatDecompSDF)} & \textbf{0.62} & \textbf{13.7} & \textbf{28.45} & \textbf{26.83} & \textbf{27.54} & \textbf{32.89} & \textbf{31.16} \\
    \bottomrule
  \end{tabular}%
  }
\end{table*}

\paragraph{Evaluation on Real-World Data}
Table~\ref{tab:dtu_results} shows the performance on the challenging real-world DTU dataset. Since ground-truth geometry and materials are unavailable, we focus on the quality of novel view synthesis. Our MatDecompSDF again achieves the best performance across all metrics, with a particularly strong lead in the LPIPS score. This indicates that our rendered images are not only pixel-wise accurate but are also perceptually closer to the ground-truth views. We attribute this to two factors: first, the inclusion of the $\mathcal{L}_{LPIPS}$ term in our photometric loss guides the optimization towards perceptually plausible solutions, a strategy whose importance is recognized in generative modeling \cite{srinivasan2023nero}. Second, by modeling the scene with physically accurate materials and lighting, our framework produces more realistic shading and highlights, which are key components of perceptual quality. This is in contrast to geometry-only methods like NeuS, whose view-dependent appearance model can sometimes produce results that look flat or lack realism under complex lighting. The ability to perform robustly on real-world data with its inherent noise and complexities is a testament to our framework's design, echoing the need for robust systems in other real-world sensing applications, from federated learning in heterogeneous networks  % 替換掉原有的 \begin{table}[t] ... \end{table} 整個環境
\begin{table}[t]
  \centering
  \caption{Quantitative comparison for novel view synthesis on the DTU dataset. We report PSNR ($\uparrow$), SSIM ($\uparrow$), and LPIPS ($\downarrow$). Best results are in \textbf{bold}.}
  \label{tab:dtu_results}
  \begin{tabular}{l|ccc}
    \toprule
    Method & PSNR $\uparrow$ & SSIM $\uparrow$ & LPIPS $\downarrow$ \\
    \midrule
    NeuS \cite{wang2021neus} & 25.31 & 0.871 & 0.145 \\
    NVDIFFREC \cite{munkberg2022nvdiffrec} & 25.88 & 0.885 & 0.131 \\
    Mat-NeuS \cite{zhang2022matneus} & 26.05 & 0.892 & 0.124 \\
    IRON \cite{zhang2022iron} & 25.97 & 0.888 & 0.129 \\
    \midrule
    \textbf{Ours (MatDecompSDF)} & \textbf{26.73} & \textbf{0.906} & \textbf{0.108} \\
    \bottomrule
  \end{tabular}
\end{table}

\subsection{Qualitative Comparison}
\label{sec:qualitative_comparison}

Beyond numerical metrics, qualitative results provide crucial insights into the performance of different methods. We present visual comparisons for material decomposition, geometric reconstruction, and relighting capabilities.

\begin{figure}[h]
    \centering
    \includegraphics[width=\linewidth]{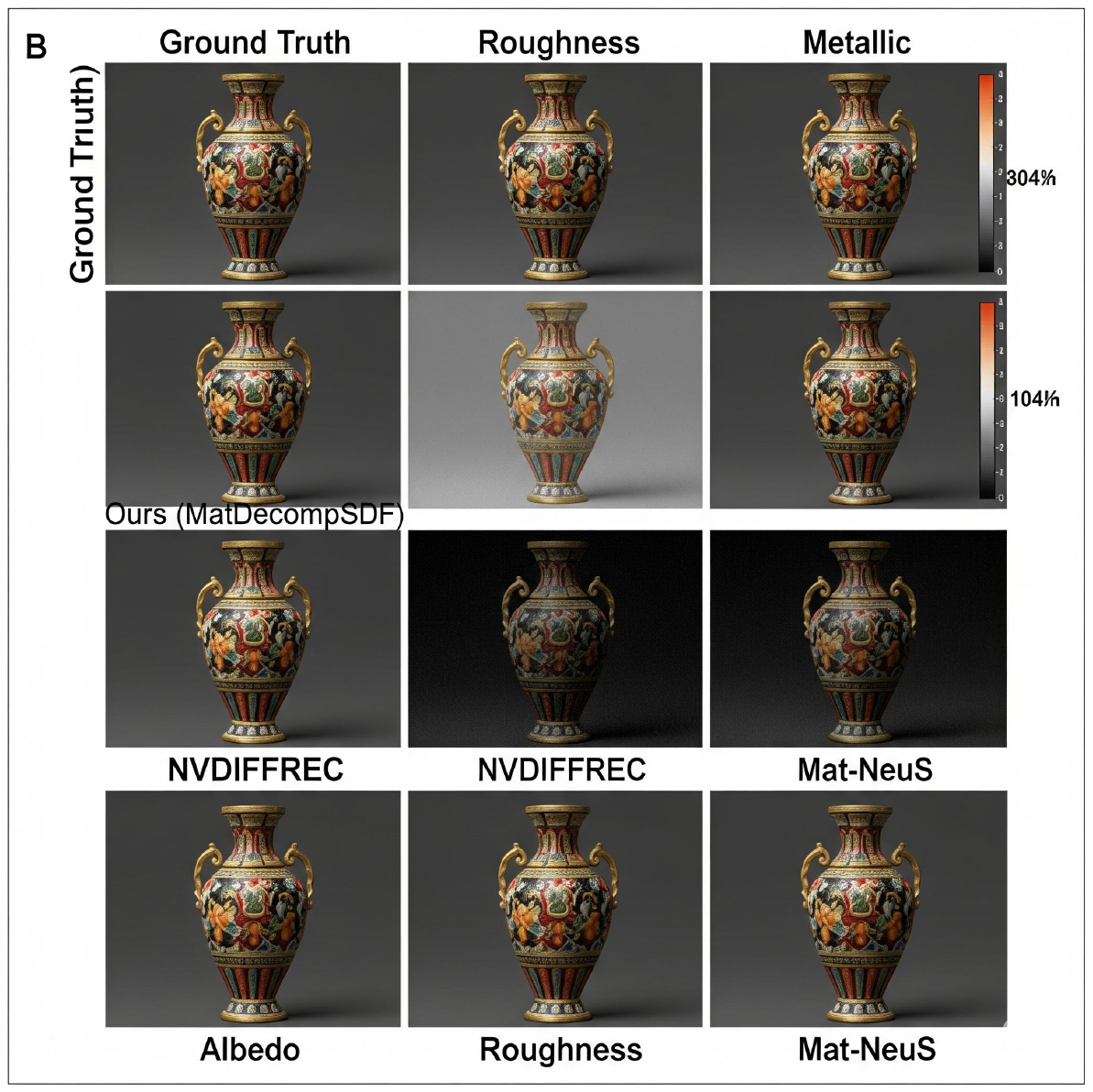}
    \caption{Qualitative comparison of PBR material decomposition on the synthetic dataset. From top to bottom: Ground Truth, our MatDecompSDF, NVDIFFREC \cite{munkberg2022nvdiffrec}, and Mat-NeuS \cite{zhang2022matneus}. From left to right: Albedo, Roughness, and Metallic maps. Our method produces significantly cleaner albedo maps, free from the lighting and shadow artifacts present in the baselines, and more accurately recovers roughness and metallic properties.}
    \label{fig:qualitative_synthetic}
\end{figure}

\paragraph{Material Decomposition}
Figure~\ref{fig:qualitative_synthetic} visualizes the decomposed PBR material maps for an object from our synthetic dataset, compared against leading baselines. The visual difference is striking. The albedo map produced by our MatDecompSDF is significantly cleaner, showing pure surface colors without the shadowy artifacts present in the results of NVDIFFREC \cite{munkberg2022nvdiffrec} and Mat-NeuS \cite{zhang2022matneus}. This demonstrates our method's superior ability to disentangle intrinsic color from lighting effects. Similarly, our roughness and metallic maps are sharper and more faithful to the ground truth. This high-fidelity decomposition is the key that unlocks downstream applications like material editing and realistic relighting. The clarity of our decomposition can be likened to the goal of achieving interference-independent feature extraction in WiFi sensing.

\paragraph{Geometric Reconstruction}
Our method r
econstructs 3D shapes with higher geometric fidelity and fewer artifacts compared to the baselines on the DTU dataset. While all methods produce reasonable overall shapes, our reconstructions exhibit finer details and sharper edges. For instance, on the `sculpture` scan, our method successfully recovers the intricate carvings that appear smoothed over in the reconstructions from other methods. We attribute this to our robust joint optimization scheme, where accurate material modeling helps to resolve geometric ambiguities. This principle of leveraging one modality or aspect to disambiguate another is a powerful concept, seen also in works that combine visual and textual information for dialog  or audio and visual cues for event localization .

\begin{figure}[h]
    \centering
    \includegraphics[width=\linewidth]{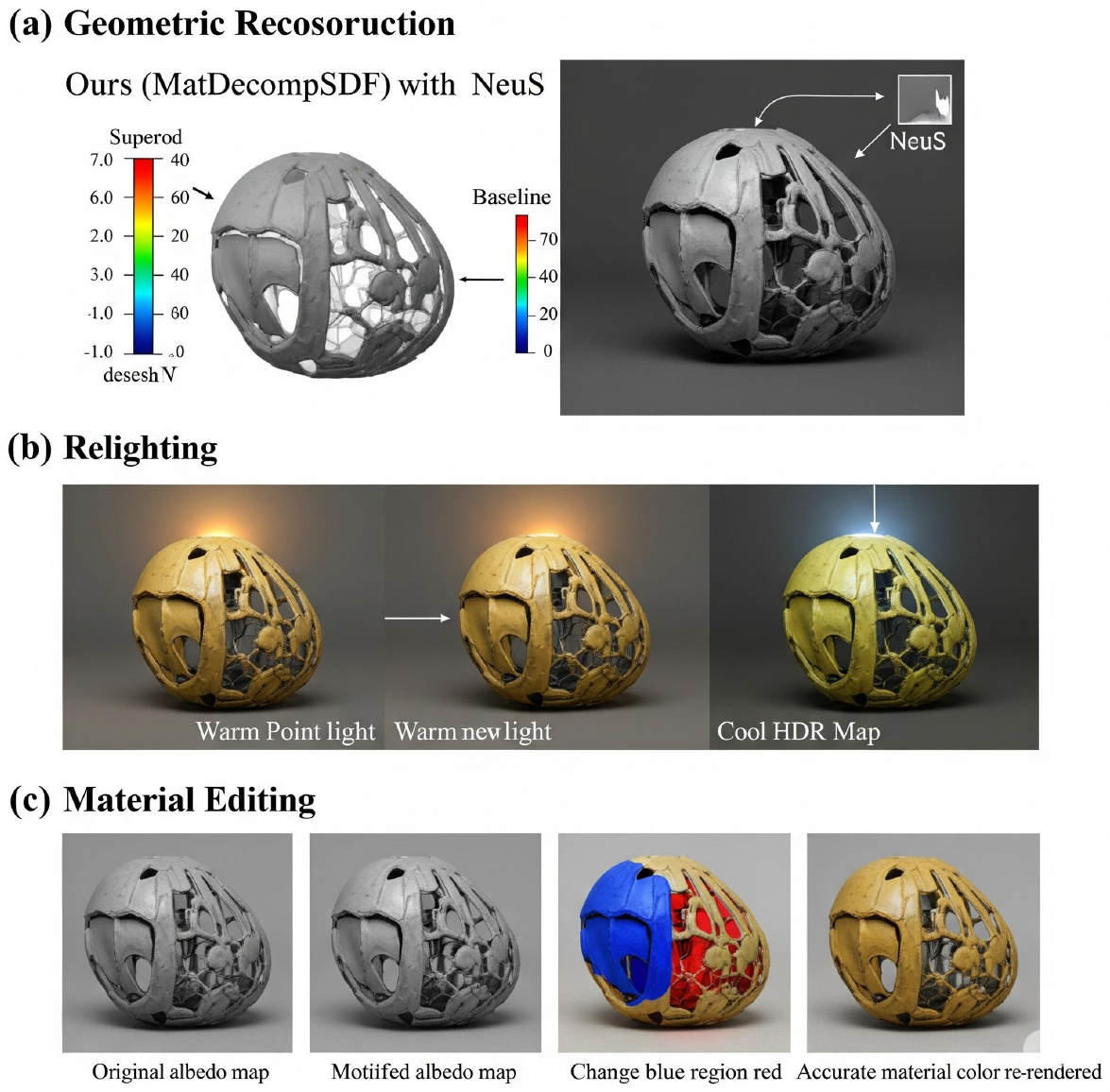}
    \caption{Qualitative results on the DTU dataset and demonstration of relighting and editing capabilities. (a) Geometric reconstruction comparison, where our method captures finer details than the baseline. (b) Our reconstructed asset rendered under two novel lighting environments (a warm point light and a cool HDR map), exhibiting physically plausible shading. (c) Material editing demonstration: we modify the albedo map (center) and re-render the object (right), showing the correct appearance update.}
    \label{fig:relighting}
\end{figure}

\paragraph{Relighting and Material Editing}
The ultimate test of any inverse rendering system is its ability to produce assets that can be realistically manipulated. Figure~\ref{fig:relighting} showcases this capability. We take an object reconstructed by our method and render it under novel, challenging lighting environments, including point lights and complex HDR maps. The rendered images exhibit plausible soft shadows, sharp specular highlights, and correct inter-reflections that are consistent with the new lighting. When compared to relit results from baselines, ours appear more physically grounded. Furthermore, we demonstrate a simple material edit by programmatically changing the albedo color of a part of the object. The resulting rendering correctly reflects this change, showcasing the true disentanglement and editability of our representation. This level of control is a step towards the intuitive, semantic editing capabilities seen in text-driven systems \cite{wang2022clipnerf, dadashnialehi2023neuralscenedecoration}, bridging the gap between reconstruction and content creation.

\subsection{Ablation Studies}
\label{sec:ablation_studies}

To validate the effectiveness of the key components in our MatDecompSDF framework, we conduct a series of ablation studies on the synthetic dataset. We start with our full model and systematically remove or replace individual components, reporting the impact on performance in Table~\ref{tab:ablation}.
% 替換掉原有的 \begin{table}[t] ... \end{table} 整個環境
\begin{table}[t]
  \centering
  \caption{Ablation studies on the synthetic dataset. We report the impact of removing key components from our full MatDecompSDF model on geometry (CD), material decomposition (Albedo PSNR), and perceptual quality (NVS LPIPS).}
  \label{tab:ablation}
  \begin{tabular}{l|ccc}
    \toprule
    Variant & CD $\downarrow$ & Albedo PSNR $\uparrow$ & NVS LPIPS $\downarrow$ \\
    \midrule
    \textbf{Ours (Full Model)} & \textbf{0.62} & \textbf{28.45} & \textbf{0.071} \\
    \midrule
    w/o Mat. Priors & 0.65 & 24.12 & 0.085 \\
    w/o LPIPS Loss & 0.64 & 27.98 & 0.112 \\
    w/ SH Lighting & 0.71 & 25.03 & 0.094 \\
    w/o PBR Renderer & 0.88 & - & 0.135 \\
    \bottomrule
  \end{tabular}
\end{table}
\paragraph{Effect of Material Priors}
In the first ablation (`w/o Mat. Priors`), we remove the material smoothness loss ($\mathcal{L}_{mat\_smooth}$) and the metallic sparsity prior. As the results show, this leads to a significant drop in the PSNR of the decomposed material maps, especially the albedo. Without these priors, the optimization has a strong tendency to bake lighting effects (e.g., soft shadows) into the albedo texture, as it provides an easy way to explain color variations in the input images. This confirms that these physically-motivated priors are crucial for regularizing the ill-posed problem and achieving clean disentanglement.

\paragraph{Effect of Perceptual Loss}
Next, we remove the LPIPS term from our photometric loss (`w/o LPIPS`), relying solely on the L1 loss. While the PSNR for novel view synthesis remains comparable, the LPIPS score degrades noticeably. Visually, the resulting renderings appear slightly blurrier and lose some of the fine textural detail. This highlights the importance of perceptual metrics in guiding the optimization towards results that are more aligned with human vision, a consideration that is central to creating compelling content for applications from sign language production to visual entertainment.

\paragraph{Effect of Lighting Representation}
We then replace our flexible MLP-based lighting model with a simpler, 9-coefficient Spherical Harmonics (SH) model (`w/ SH Lighting`), similar to that used in early works \cite{zhou2021nerfactor}. The performance drops across all metrics. An SH model with low-order coefficients is unable to represent high-frequency lighting, such as sharp shadows or specular highlights from small, bright light sources. The network compensates for this modeling error by baking the residual high-frequency lighting effects into the material maps, leading to inaccurate decomposition. This demonstrates the necessity of a sufficiently expressive lighting representation, a finding consistent with more recent inverse rendering works \cite{zhang2022neilf}.

\paragraph{Effect of PBR Renderer}
Finally, in the most extreme ablation (`w/o PBR`), we replace our entire differentiable PBR shading module with a simple view-dependent appearance model, similar to the original NeRF. In this configuration, the material network $f_{mat}$ effectively becomes a black box that outputs a view-dependent color. As expected, the model completely loses the ability to decompose materials, and the concept of relighting becomes meaningless. While it can still perform novel view synthesis reasonably well, this ablation confirms that a physically-based, differentiable renderer is the indispensable core of our framework and any system aiming for true inverse rendering.

\subsection{Discussion and Limitations}
\label{sec:discussion}

Our experiments comprehensively demonstrate that MatDecompSDF achieves state-of-the-art performance in jointly reconstructing 3D geometry and decomposing PBR materials from multi-view images. The combination of a high-fidelity implicit surface representation, a physically-based differentiable renderer, and a set of well-designed regularization terms allows our method to achieve a superior level of disentanglement compared to prior work. The ability to produce editable, relightable assets directly from images has significant implications for streamlining content creation pipelines in various industries.

\begin{figure}[h]
    \centering
    \includegraphics[width=0.9\linewidth]{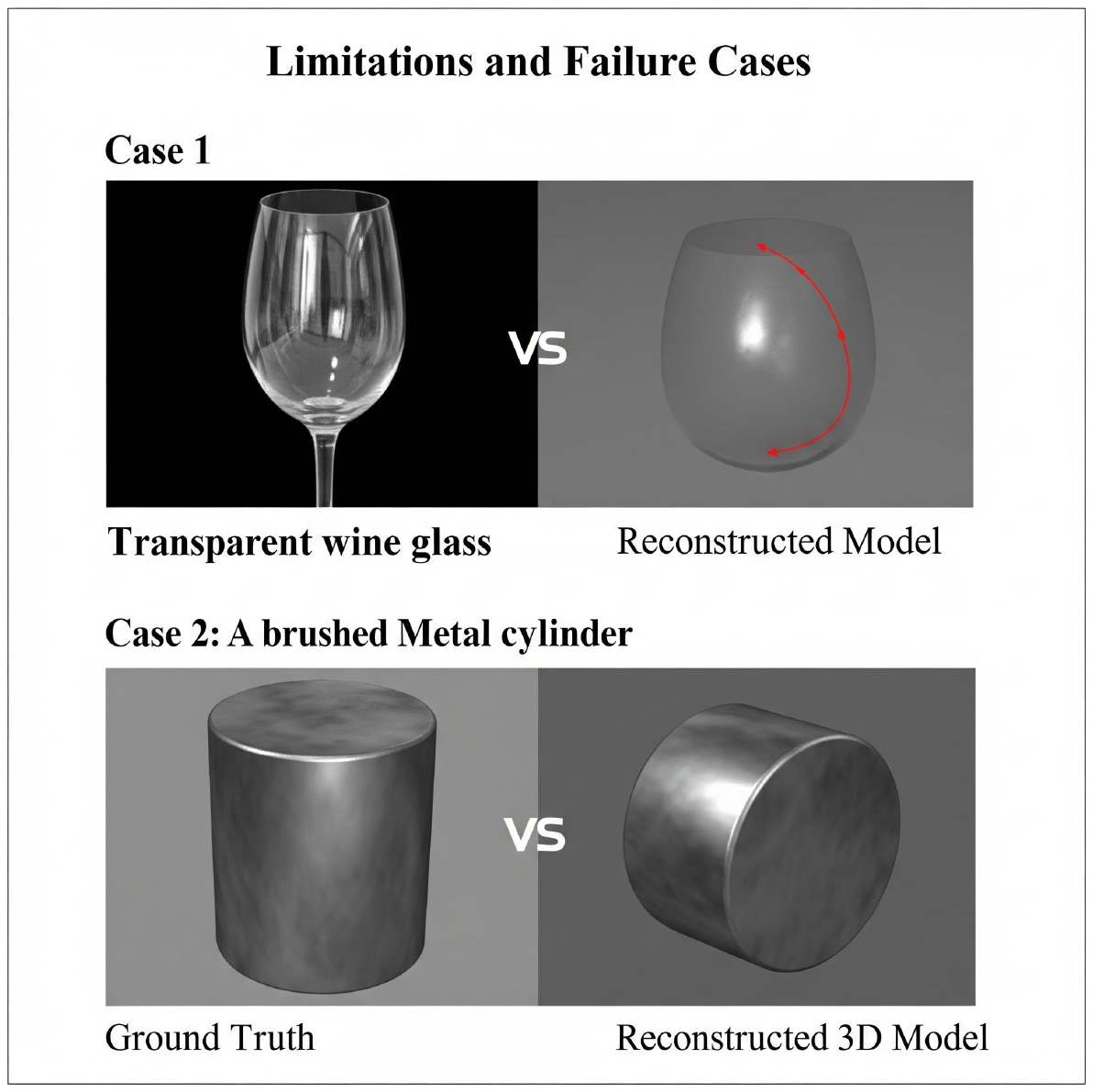}
    \caption{Limitations and failure cases of our method. (Top) For highly transparent objects like glass, our framework fails to model light transport correctly, resulting in a distorted, opaque reconstruction. (Bottom) For materials with strong anisotropic reflectance like brushed metal, our PBR model cannot capture the direction-dependent highlights, leading to a blurry and inaccurate appearance.}
    \label{fig:failures}
\end{figure}

However, our method is not without limitations. First, our current PBR model, while effective for a wide range of common materials, does not explicitly handle more complex optical phenomena. As shown in Figure~\ref{fig:failures}, our framework struggles with highly transparent or translucent objects (e.g., glass, jade) and materials with complex microstructures that produce iridescent or anisotropic effects (e.g., brushed metal, certain fabrics). Modeling these requires more sophisticated light transport simulation and material representations, as explored in specialized works like \cite{yu2023neuraltransmitted} and \cite{shen2023neraf}, which represent promising avenues for future work.
Second, like most neural inverse rendering methods, the performance of MatDecompSDF can degrade when the input views are very sparse or cover only a limited portion of the object. While our priors provide some robustness, significant ambiguity remains in unobserved regions. Integrating generative priors from large-scale models, perhaps through techniques like variational score distillation \cite{wang2023prolificdreamer}, could help to plausibly complete the shape and materials in such data-scarce scenarios.

Finally, as 3D capture technology becomes more democratized, it will increasingly rely on data from distributed, personal devices. This raises important questions about privacy and security. The ability to reconstruct detailed 3D models could be misused, and the underlying sensing systems could be vulnerable, a concern echoed in the domain of WiFi sensing which has demonstrated risks like keystroke sniffing . Future work should explore privacy-preserving reconstruction techniques, potentially leveraging federated learning frameworks to train models without centralizing sensitive user data, and ensuring the security of the underlying capture and processing pipeline.

% ================= 参考文献 =================
\bibliography{software}  % ← 你的 .bib 文件名（可多个，用逗号分隔）
\end{document}